\documentclass{article} %
\usepackage{iclr2026_conference,times}

\usepackage{amsmath,amsfonts,bm}

\def\eqref#1{equation~\ref{#1}}

\def\1{\bm{1}}

\DeclareMathAlphabet{\mathsfit}{\encodingdefault}{\sfdefault}{m}{sl}
\SetMathAlphabet{\mathsfit}{bold}{\encodingdefault}{\sfdefault}{bx}{n}

\usepackage{hyperref}
\usepackage{url}
\usepackage{xspace}
\usepackage{graphicx}
\usepackage{booktabs}
\usepackage{tabularx}
\usepackage{wrapfig}
\usepackage[T1]{fontenc}
\usepackage{pifont}
\usepackage{subcaption}

\newcommand{\method}{SpAx\xspace}

\newif\ifincludesupplement

\includesupplementtrue %

\title{\raggedright Activation Sparsity with Weight Approximation for Faster LLM Decoding on Offloaded Weights}

\author{JuneHyung Kim$^{1}$\thanks{Work done while at University of Toronto.}, \ Sankeerth Durvasula$^{2}$, \ Nandita Vijaykumar$^{2}$ \\
$^{1}$LG Electronics \quad $^{2}$University of Toronto \\
\texttt{junehyung7.kim@lge.com}, \ \texttt{\{sankeerth,nandita\}@cs.toronto.edu}
}

\iclrfinalcopy
\begin{document}

\maketitle
\lhead{Preprint}

\begin{abstract}
Deploying LLMs on consumer-grade GPUs with insufficient memory to hold their weights can result in prohibitively slow inference, because decoding repeatedly transfers offloaded weights from system RAM or flash storage into GPU at much lower bandwidth than local GPU-memory access. Activation sparsity reduces these transfers by skipping weights associated with zero or near-zero activations. However, as more activation contributions are omitted, model quality eventually degrades rapidly, indicating that weights associated with small-magnitude activations collectively influence model quality sharply. In this work, we improve the trade-off between model quality and decoding performance when exploiting activation sparsity. Our key idea is to replace the binary choice of whether or not to read a weight with three options: fully retain it, approximate it using a compressed weight representation, or omit it entirely. \method skips weights associated with activations closest to zero, reads approximate weights for smaller-magnitude activations, and reads original weights for the largest-magnitude activations. Smaller-magnitude activations attenuate the errors introduced by approximate weights, while compressed weight representations require fewer bytes to be transferred. With weights offloaded to CPU memory, \method speeds up decoding by $3.86\times$ on average (up to $5.57\times$) with 16-bit weights and $2.06\times$ (up to $2.74\times$) with 4-bit weights, at a WikiText-2 perplexity increase of at most $10\%$. With weights offloaded to flash storage, the speedups are $3.31\times$ on average (up to $4.81\times$) and $1.54\times$ (up to $2.03\times$).
\end{abstract}

\section{Introduction}
\label{sec:intro}

Large language models (LLMs) have demonstrated strong capabilities in a wide range of tasks, from text understanding to agentic workflows such as code generation, computer use, and mathematical reasoning. Running capable LLMs locally on personal devices such as desktops or workstations gives users greater control over their data and computing. This has vital implications for task ownership and data privacy, and reduce dependence on cloud subscriptions and their dynamic pricing.

Unlike server-grade GPUs, desktop and workstation GPUs typically have limited GPU memory. We use \emph{GPU memory} for VRAM, \emph{CPU memory} for system DRAM, and \emph{flash storage} for SSD-based storage (an NVMe SSD in our experiments). \emph{Offloaded weights} reside in CPU memory or flash storage and are transferred into GPU memory as needed during decoding. These transfers are much slower than reading weights already resident in GPU memory. For example, on our workstation (Section~\ref{sec:results:setup}), the NVIDIA RTX 4090's local memory bandwidth is about $80\times$ the bandwidth for transferring weights from CPU memory and about $150\times$ that for transferring weights from flash storage.
Thus, transferring offloaded weights can dominate decoding latency~\citep{alizadeh2024flash,xue2024powerinfer2}. Reducing this latency requires reducing the weight bytes transferred to generate each token. This requires targeting the transformer's attention projection and feed-forward network (FFN) weights. Within each projection, a \emph{weight column} contains the weights multiplied by a single input activation entry (see Section~\ref{sec:background:sparsity}). Activation sparsity reduces these transfers by skipping weight columns associated with zero activations. The columns that can be skipped differ across tokens.

Early methods such as Deja Vu~\citep{liu2023dejavu} and PowerInfer~\citep{song2024powerinfer} demonstrate that activation-dependent sparsity can accelerate LLM inference. These methods exploit sparse activations in the FFN of models that use the ReLU activation function~\citep{nair2010rectified}. However, this source of sparsity is largely absent in modern LLMs that use smooth activation functions such as SiLU and GELU~\citep{shazeer2020glu}. ReLUfication~\citep{mirzadeh2024relu,song2025prosparse,song2024turbosparse} addresses this limitation by replacing the smooth activation function with ReLU and performing continued pretraining to recover model quality. PowerInfer-2~\citep{xue2024powerinfer2} decodes such sparsified models on a smartphone with FFN weights offloaded to flash storage. Continued pretraining, however, requires access to representative training data and compute resources and produces a modified instance of the model.
Training-free approaches instead induce sparsity by omitting reads of weight columns corresponding to low-magnitude activations. CATS~\citep{lee2024cats} applies this approach to FFN weights in models using SiLU activation functions. CHESS~\citep{he2024chess} and TEAL~\citep{liu2025teal} apply magnitude-based sparsification to attention projection weights as well. These methods offer a sparsity level hyperparameter that is adjustable during deployment, allowing a trade-off between model quality and weights read per token. Increasing sparsity removes contributions of progressively higher-magnitude activations. However, beyond a certain sparsity level, even a small increase in leveraged sparsity causes model quality to drop sharply (see Section~\ref{sec:motivation}). This rapid degradation indicates that model quality is highly sensitive to omitting these intermediate-magnitude activations. This limits the sparsity level that can be used in practice~\citep{szatkowski2026universal}.

Our goal in this work is to accelerate autoregressive decoding in modern LLMs on consumer-grade GPUs where weights exceed GPU memory, without requiring continued pretraining. To this end, we propose \method (\textbf{Sp}arsity with \textbf{A}ppro\textbf{x}imation), a tiered activation-sparsity mechanism to reduce the \emph{weight data} required per token, i.e., the number of weight bytes (compressed or otherwise) transferred to GPU memory, while limiting model-quality loss.
The key idea of \method is to replace the binary choice of whether or not to read a weight column with three options: read the original weights of the weight column, read an approximation of the weight column using a compressed representation, or omit it entirely. This design builds on the observation that an activation's magnitude scales the contribution of its weight column's approximation error to the projection's output error. Because smaller-magnitude activations attenuate these errors, approximating their weight columns with compressed representations loses little accuracy while transferring far fewer bytes. This motivates omitting weights associated with activations closest to zero, approximating those of smaller-magnitude activations, and retaining the original weights for the largest-magnitude activations.

Realizing this idea requires addressing three challenges:
(1) providing compressed weight approximations that preserve useful contributions and allow each selected column to be read independently;
(2) choosing which columns to retain, approximate, or omit at each token under a given read budget; and
(3) translating byte savings into faster decoding despite the fragmented storage reads caused by token-dependent column selection.
To address (1), \method defines three weight approximation tiers: a top tier read in the deployment's weight format, a middle tier read from compressed representations of each weight that the GPU decompresses during decoding, and a bottom tier left unread. The middle tier encodes each weight column independently (Section~\ref{sec:method:codes}).
To address (2), \method retains the magnitude ordering used by top-$k$ selection and introduces a second boundary to divide activation entries among the three tiers. Offline calibration allocates the read budget among matrices and between the top and middle tiers (Section~\ref{sec:method:calibration}).
To address (3), \method groups matrices that share input activations and places their jointly accessed weight columns contiguously in storage. It merges adjacent selected ranges into larger read requests to reduce fragmentation (Section~\ref{sec:method:streaming}).

We evaluate \method on three models in two weight formats against four training-free activation-sparsity methods. 
At a WikiText-2 perplexity increase of at most $10\%$, with weights offloaded to CPU memory, \method decodes $3.86\times$ faster than dense on average (up to $5.57\times$) in BF16 and $2.06\times$ (up to $2.74\times$) with 4-bit weights, where TEAL reaches $1.94\times$ and $1.85\times$ on average.
With flash offloading, the speedups are $3.31\times$ (up to $4.81\times$) and $1.54\times$ (up to $2.03\times$), where TEAL reaches $1.82\times$ and $1.48\times$.
Our contributions are as follows:
\begingroup
\setlength{\leftmargini}{1.25em}
\setlength{\topsep}{2pt}
\setlength{\partopsep}{0pt}
\setlength{\parskip}{0pt}
\begin{itemize}
    \item We propose \method, a training-free method that leverages activation-sparsity to selectively read each offloaded weight in full precision, as compressed approximations, or not read at all, according to its activation's magnitude.
    \item At the same model quality, \method requires significantly less weight data per token than prior work in all evaluated settings.
    \item \method supports offloading to CPU memory or flash storage, with 16- or 4-bit weights.
\end{itemize}
\endgroup

\vspace{-6pt}
\section{Background}
\label{sec:background}
\label{sec:background:sparsity}

\vspace{-6pt}

\begingroup
\setlength{\abovedisplayskip}{4pt}
\setlength{\belowdisplayskip}{4pt}
\setlength{\abovedisplayshortskip}{2pt}
\setlength{\belowdisplayshortskip}{2pt}

\paragraph{Weights of a Transformer Layer}
A gated feed-forward network (FFN; SwiGLU or GeGLU~\citep{shazeer2020glu}) computes an intermediate activation $a = \sigma(W_{\mathrm{gate}}\,x) \odot W_{\mathrm{up}}\,x$ from its input $x$, where $\sigma$ is SiLU or GELU, and projects it back to the model dimension as $\mathrm{FFN}(x) = W_{\mathrm{down}}\,a$.
An attention block~\citep{vaswani2017attention} projects its input $h$ into queries, keys, and values with $W_q$, $W_k$, and $W_v$, and projects the concatenated head outputs $z$ back to the model dimension with $W_o$.
\method is applied to each of these seven matrices. Their inputs are $h$ for $W_q$, $W_k$, and $W_v$, $z$ for $W_o$, $x$ for $W_{\mathrm{gate}}$ and $W_{\mathrm{up}}$, and $a$ for $W_{\mathrm{down}}$.

\paragraph{Activation sparsity: omitting columns of weight matrices}
For any of these FFN and attention projection weights, let $W\in\mathbb{R}^{d\times n}$ be its weight matrix and $x\in\mathbb{R}^{n}$ its input activation. The product $Wx=\sum_{i=1}^{n}x_iW_{:,i}$ scales each \emph{weight column} $W_{:,i}$ by its activation entry $x_i$ and sums the results.
Activation sparsity computes this product from a subset of the entries, so only their weight columns need to be transferred. In a ReLU FFN, exact-zero entries can be skipped without changing the output. Gated FFNs with smooth activation functions produce almost no exact zeros, so activation sparsity instead omits entries with small magnitudes~\citep{lee2024cats,liu2025teal}.
A binary mask $m\in\{0,1\}^{n}$ selects the entries to retain:
\begin{equation}
W(m\odot x) = \sum_{i:m_i=1} x_i W_{:,i}.
\label{eq:background:sparse_projection}
\end{equation}
Each omitted entry $x_i$ adds an error $e_i$ to the output:
\begin{equation}
e_i = x_i W_{:,i}, \qquad \|e_i\|_2 = |x_i|\,\|W_{:,i}\|_2, \qquad
Wx - W(m\odot x) = \sum_{i:m_i=0} e_i.
\label{eq:background:omission_error}
\end{equation}
The error terms scale with the activation entry's magnitude, while the total error also depends on the directions of the omitted weight columns.
With at most $k$ retained entries, the best mask is:
\begin{equation}
m^{\star}\in
\underset{m\in\{0,1\}^{n},\ \|m\|_0\leq k}{\arg\min}
\left\|Wx-W(m\odot x)\right\|_2^2.
\label{eq:background:sparsity_objective}
\end{equation}

A simple selection rule that works well in practice (though not optimal) is magnitude-based top-$k$, which retains the $k$ entries with the largest $|x_i|$ and gives a sparsity level of $1-k/n$. Top-$k$ reduces the error due to discarded activations $\sum_{i:m_i=0} x_i^2$. 

\endgroup

\vspace{-6pt}
\section{Related Work}
\label{sec:related_work}
\vspace{-6pt}

\textbf{Training-free activation sparsity.} These methods reduce the weight data required during decoding without updating model parameters through fine-tuning or continued pretraining. 
CATS~\citep{lee2024cats} omit loading weights of small-magnitude activations following the FFN's gate projection activation function. TEAL~\citep{liu2025teal} extends this technique to every weight matrix in the model.
WINA~\citep{chen2026wina} and LaRoSA~\citep{liu2025larosa} refine the selection criterion: WINA scores an entry by its magnitude and the norm of the weight column it indexes, while LaRoSA ranks entries in a rotated activation vector. 
R-Sparse~\citep{zhang2025rsparse} multiplies the largest-magnitude entries by the original weights and the remaining entries by a rank-$r$ SVD approximation of the weight matrix.
We compare \method with TEAL, LaRoSA, WINA and R-Sparse in Section~\ref{sec:results}.

\textbf{Training-based activation sparsity.} Training-based approaches modify pretrained model weights to promote higher sparsity in input activations. ReLUfication~\citep{mirzadeh2024relu,song2024turbosparse} replaces smooth activation functions with ReLU variants and uses continued pretraining to recover model quality. 
Systems such as Deja Vu~\citep{liu2023dejavu} and PowerInfer~\citep{song2024powerinfer} exploit sparse activations in ReLU-based FFNs, using trained predictors to identify active neurons before computing the full activation. PowerInfer-2 uses sparsified checkpoints to enable inference on smartphones with limited memory~\citep{xue2024powerinfer2}.

\textbf{Caching methods.} Reducing the number of selected weights does not by itself ensure efficient storage access: decoding latency also depends on the locality of reads, cache reuse, and overlap between I/O and computation. LLM in a Flash~\citep{alizadeh2024flash} stores FFN weights in flash storage and fetches only the columns associated with neurons that its predictor marks active. To reduce small random reads, LLM in a Flash and Neuralink place weights accessed together next to one another~\citep{alizadeh2024flash,wang2025neuralink}. Neuron Chunking incorporates storage cost into selection, choosing contiguous neuron chunks according to their importance and estimated I/O latency~\citep{yang2025neuronchunking}. DIP's cache-aware masking considers both activation magnitude and the weights already resident in memory to reduce flash reads~\citep{federici2025dip}. ActiveFlow overlaps storage I/O with computation by predicting and preloading active weights for subsequent layers~\citep{jia2025activeflow}. \method addresses the storage access cost of its tiered reads by placing jointly accessed columns contiguously and merging adjacent selected ranges into larger requests.

\textbf{Offline weight compression.} Weight quantization reduces data transfer by storing every weight at a lower precision. AWQ~\citep{lin2024awq} uses activation magnitudes to identify salient weights and protects them through scaling.
QTIP~\citep{tseng2024qtip} codes weights as a path through a trellis instead of an index into a codebook, reaching a lower error at the same bitrate.
Activation sparsity and weight quantization can be combined \citep{liu2025teal,zhang2025rsparse,wang2025dynsparsequant}. These approaches read every retained column from a single quantized representation. \method instead stores both the deployment-format weights and a lower-bit copy, and decides at each token which columns to read from each.
Concurrent work GRINQH~\citep{oberlander2026grinqh} uses activation magnitudes at each token to select weight-column precision or skip columns. We omit direct comparison because we could not locate public code at submission time.

\vspace{-6pt}
\section{Model Quality on Omitting Weights}
\label{sec:motivation}
\vspace{-6pt}

\begin{figure}[!htb]
\centering
\includegraphics[width=0.9\textwidth]{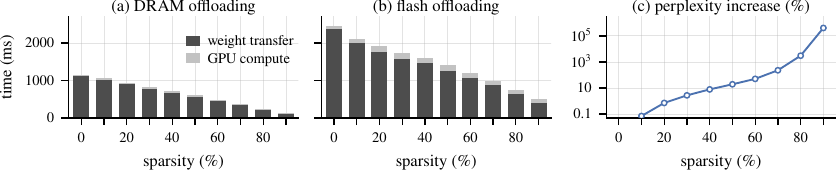}
\vspace{-8pt}
\caption{Llama-3.1-8B (BF16) with all weights offloaded, at different sparsity levels.
\textbf{(a), (b):} Time per decoded token with the weights offloaded to CPU memory and to flash storage.
\textbf{(c):} Perplexity increase relative to dense on the WikiText-2 test set.
(See Section~\ref{sec:results:setup} for the experiment setup.)}
\label{fig:dial}
\vspace{-12pt} %
\end{figure}

Figure~\ref{fig:dial} shows the speed--quality tradeoff of TEAL~\citep{liu2025teal} on Llama-3.1-8B with weights offloaded to CPU memory and flash storage. With CPU memory offloading, decoding takes $1.16$~s per token, $97\%$ of it on weight transfers. Omitting $40\%$ of columns yields a $1.62\times$ speedup with under $7.8\%$ perplexity increase; omitting $70\%$ yields a $3.17\times$ speedup but sharply raises perplexity.

As sparsity increases, top-$k$ selection omits columns associated with larger activations. Each column's omission error scales with activation magnitude (Equation~\ref{eq:background:omission_error}). AWQ likewise finds that protecting channels associated with large activations reduces quantization error~\citep{lin2024awq}. R-Sparse shows that even coarse approximations of omitted contributions can recover model quality~\citep{zhang2025rsparse}. This motivates reading compressed approximations of these columns to retain some of their contribution while transferring fewer bytes than the original weights.

\vspace{-6pt}
\section{\method: Tiered Activation Sparsity}
\vspace{-6pt}
\label{sec:method}

\label{sec:method:overview}

\begin{figure}[!t]
\centering
\includegraphics[width=\textwidth]{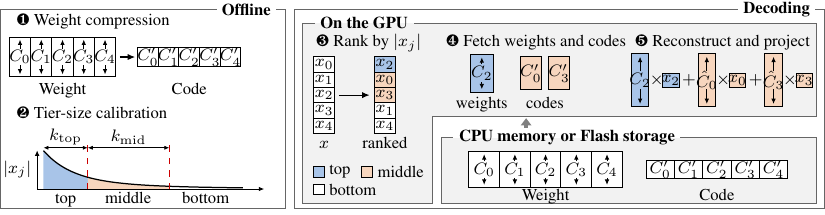}
\vspace{-8pt}
\caption{Overview of \method. Offline encoding and calibration prepare compressed weight codes and determine tier sizes. During decoding, activation magnitudes determine which columns to read in the original format, reconstruct from compressed codes, or skip.}
\label{fig:overview}
\vspace{-12pt} %
\end{figure}

\method reduces the weight data required per token by assigning weight columns to three tiers according to their activation magnitudes.
It reads the \emph{top tier} in the deployment's weight format, reads compressed approximations for the \emph{middle tier}, and leaves the \emph{bottom tier} unread.
Figure~\ref{fig:overview} illustrates the offline preparation and per-token decoding steps.
\method encodes each weight column $C_j$ into compressed codes $C'_j$ (step~1; Section~\ref{sec:method:codes}).
It then calibrates the number of columns assigned to each tier for each weight matrix under a given read budget (step~2; Section~\ref{sec:method:calibration}). The original weights and their compressed codes are stored in CPU memory or flash storage.
During decoding, at each token, \method ranks the input activations by magnitude and assigns their weight columns to the three tiers (step~3).
In the example, $x_2$ selects $C_2$ for the top tier, $x_0$ and $x_3$ select $C_0$ and $C_3$ for the middle tier, and the remaining columns belong to the bottom tier.
\method fetches $C_2$ and the compressed codes $C'_0$ and $C'_3$ into GPU memory, leaving $C_1$ and $C_4$ unread (step~4).
The GPU reconstructs $\hat{C}_0$ and $\hat{C}_3$ from these codes and computes $x_2C_2+x_0\hat{C}_0+x_3\hat{C}_3$ (step~5).

\subsection{Compressed Representation for the Middle Tier}
\label{sec:method:codes}

For the middle tier, we consider two lossy weight compression schemes: scalar quantization and trellis-coded quantization (step~1 in Figure~\ref{fig:overview}). 

\textbf{Scalar quantization.}
We represent each weight using a two-bit code and store two scaling factors, $s_{j,0}$ and $s_{j,1}$, shared by all weights in column $C_j$.
The two bits of the code specify signs $e_1,e_2\in\{-1,+1\}$, and the weight is reconstructed as $e_1s_{j,0}+e_2s_{j,1}$.
In \method, we encode all weight columns offline using symmetric two-bit Lloyd--Max quantization~\citep{lloyd1982least,max1960quantizing}.
We first determine lower and upper representative weight magnitudes, $l_j$ and $h_j$, for each column $C_j$.
We set $s_{j,0}=(h_j+l_j)/2$ and $s_{j,1}=(h_j-l_j)/2$, so that adding or subtracting the two scales yields $h_j$ or $l_j$, respectively.
We then select two signs $e_1,e_2 \in \{-1, +1\}$ for each weight $W_{ij}$ to minimize its squared reconstruction error with the scales fixed as follows:
\begin{equation}
(e_1,e_2) = \underset{(a,b)\in\{-1,+1\}^2}{\arg\min}
\left(W_{ij}-a s_{j,0}-b s_{j,1}\right)^2,
\qquad
\hat{W}_{ij}=e_1s_{j,0}+e_2s_{j,1}.
\label{eq:method:scalar_quantization}
\end{equation}

During decoding, \method reads the codes and two scales of each selected middle-tier column and reconstructs its weights on the GPU.

\textbf{Trellis-coding compression.}
Prior work in quantization has shown that jointly encoding weights across multiple columns reduces quantization error compared to compressing each weight independently~\citep{tseng2024qtip, tseng2024quip, aqlm}.
Because activation sparsity reads only a few selected columns, \method instead builds on bitshift trellis coding~\citep{mao2010trellis} to encode the weights within each column jointly, so that each column can be reconstructed from its own codes.

\begin{wrapfigure}[12]{r}{0.4\textwidth}
\vspace{-\intextsep}
\centering
\includegraphics[width=\linewidth]{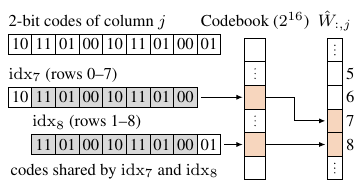}
\caption{Reconstructing a middle-tier column from its own $b$-bit codes.}
\label{fig:method}
\end{wrapfigure}
\method represents each weight column as a sequence of two-bit codes and a column-specific scale.
All columns share a codebook of $2^{16}$ values drawn once from a standard normal distribution.
Each weight is represented by a window of eight consecutive codes, its own and those of the seven preceding weights in the column, which form a 16-bit index into the codebook (Figure~\ref{fig:method}).
The selected value times the column's scale gives the approximate weight.
To encode a column, \method divides its weights by their root-mean-square and stores this value as the scale. We choose the code sequence that minimizes the total squared error between the normalized weights and their reconstructions.
Because the windows overlap, each code affects up to eight weights, so choosing each code separately need not minimize the column's error.
\method therefore chooses the codes jointly with the Viterbi algorithm~\citep{forney1973viterbi}.
During decoding, consecutive weights share seven of their eight codes. Moving to the next row drops the oldest code and appends the next one (row 7 uses the codes of rows 0--7, row 8 those of rows 1--8); codes before the start of the column are treated as zeros.
Each weight thus needs one codebook lookup and one multiplication.

\vspace{-0.6pt}
\subsection{Calibrating the Number of Weight Columns in Each Tier}
\label{sec:method:calibration}
\vspace{-0.6pt}

At each token, \method ranks the input activation entries by magnitude and splits the corresponding weight columns into the three tiers (step~3 in Figure~\ref{fig:overview}). The columns of the largest entries form the top tier ($k_{\mathrm{top}}$ in number), those of the next entries form the middle tier ($k_{\mathrm{mid}}$ in number), and the rest form the bottom tier. \method calibrates the number $k_{\mathrm{top}}$ and $k_{\mathrm{mid}}$ offline within a given read budget (step~2 in Figure~\ref{fig:overview}).
Different choices of $k_{\mathrm{top}}$ and $k_{\mathrm{mid}}$ affect both the weight data required and model quality. We express the per-matrix \emph{read budget} as the weight data required to compute $Wx$, normalized by the size of the full weight matrix $W$:
\vspace{-0.6pt}
\begin{equation}
\mathbf{budget}=\frac{k_{\mathrm{top}}+\rho\,k_{\mathrm{mid}}}{n},
\label{eq:method:budget}
\end{equation}
\vspace{-0.6pt}
where $n$ is the number of columns of $W$ and $\rho$ is the compression ratio of the middle tier. Searching $k_{\mathrm{top}}$ and $k_{\mathrm{mid}}$ for every weight matrix creates a large search space, so \method splits the search into two steps: For a given overall budget, Step~1 searches for read budget of each group of weight matrices that share an input activation, and Step~2 splits each group's budget between the top and middle tiers.
Both steps minimize the output error of Equation~\ref{eq:background:sparsity_objective} (Section~\ref{sec:background:sparsity}), averaged over a calibration dataset.

\textbf{Step 1. Greedy search for each group's budget.} Weight matrices that share an input activation share its ranking, so \method gives them one budget (Equation~\ref{eq:method:budget}, normalized by their combined dense size) and one pair of tier sizes. Each layer thus has four groups: $(W_q,W_k,W_v)$, $(W_o)$, $(W_{\mathrm{gate}},W_{\mathrm{up}})$, and $(W_{\mathrm{down}})$. Starting with every group dense (budget $1$) and $k_{\mathrm{mid}}=0$, \method iteratively searches for the next group to reduce. In each iteration, it tries lowering each group's budget by $0.05$ while holding the others fixed, and applies the reduction with the smallest $\ell_2$ output error (Equation~\ref{eq:background:sparsity_objective}) on a calibration dataset. The search stops when the layer's read budget reaches its target; the intermediate steps give the group budgets for larger targets.

\textbf{Step 2. Grid search for $k_{\mathrm{top}}$ and $k_{\mathrm{mid}}$.} With the group budgets from Step~1 fixed, each column removed from the top tier pays for $1/\rho$ middle-tier columns (Equation~\ref{eq:method:budget}). From $k_{\mathrm{mid}}=0$, the grid lowers $k_{\mathrm{top}}$ in equal steps, raising $k_{\mathrm{mid}}$ to keep the budget fixed, until the bottom or the top tier is empty. \method selects the setting with the smallest output error (Appendix~\ref{app:fractions}).

\vspace{-0.6pt}

\subsection{Efficiently Loading Offloaded Weights}
\label{sec:method:streaming}
\vspace{-0.6pt}

\begin{figure}[!htb]
\centering
\includegraphics[width=0.5\linewidth]{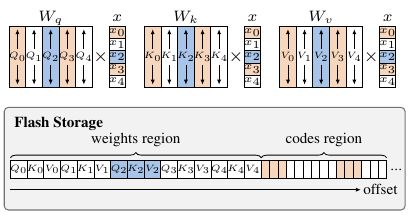}
\caption{Weight columns of $W_q$, $W_k$ and $W_v$ arranged contiguously in flash storage.}
\label{fig:method:storage}
\end{figure}

When weights are offloaded to flash storage, reads proceed at a coarse granularity and large read requests are served faster than many small ones, but token-dependent column selection fragments each token's read into many small ranges. Middle-tier columns are read in even smaller requests, because their codes occupy fewer bytes than their weights. \method therefore stores the columns of each group contiguously in the order of their activation entries and merges adjacent selected ranges into single read requests.

\method fetches the weight columns one
activation entry indexes in fewer and larger requests, by storing them in
contiguous ranges of flash storage.
It stores the codes of those columns the same way, in a separate region.
To this end, \method uses the four groups of
Section~\ref{sec:method:calibration}, whose matrices share an input activation.
An entry of that activation indexes one column of every matrix in its group, as
entry $x_2$ indexes $Q_2$, $K_2$ and $V_2$ in Figure~\ref{fig:method:storage}.
To place the columns of one entry in the same tier, \method ranks that
activation once for the group, and the two searches of
Section~\ref{sec:method:calibration} give the group one budget and one setting
$(k_{\mathrm{top}}, k_{\mathrm{mid}})$.
In the figure, $x_2$ is in the top tier, so a single request reads $Q_2$, $K_2$
and $V_2$.
If each matrix occupied its own region, the same three columns would take three
requests.
Entries $x_0$ and $x_3$ are in the middle tier, and the codes of each entry's
three columns take one request.
Each region holds its ranges in the index order of the entries, which is fixed across tokens.
At each decoding step, \method merges the adjacent ranges it reads, so a run of
neighbouring entries in one tier takes one request instead of one each.

\vspace{-0.6pt}
\section{Results}
\label{sec:results}
\vspace{-0.6pt}

\paragraph{Experiment setup}
\label{sec:results:setup}

We implement \method using llama.cpp \citep{gerganov2023llamacpp} and decode at batch one on a workstation with an NVIDIA RTX 4090 GPU. We evaluate two settings: (i) \emph{CPU memory offloading}, with weights and \method's codes held in CPU memory and transferred into GPU memory over PCIe, and (ii) \emph{flash offloading}, with the weights and codes read from flash storage (an NVMe SSD). We define the \emph{read budget} as the weight data required per decoded token, normalized by the size of the same weight matrices in the dense deployment format. This includes selected original weights, compressed codes, and any auxiliary data required by the compression format. Actual bytes transferred from flash storage can be larger because reads occur at coarser granularity.

We offload all weights targeted by each configuration to study the transfer bottleneck independently of GPU memory capacity. On our PCIe 4.0 x8 workstation, transfers account for $97\%$ of dense LLAMA-ALL-BF16 decoding time with CPU memory offloading. Desktops with smaller-memory GPUs, such as an RTX 4060, face a similar transfer bottleneck, although differences in GPU computation may change the exact speedup. CA illustrates partial residency by keeping attention projection weights in GPU memory.
Table~\ref{tab:setup} lists the models, configurations and weight formats we evaluate. Each case is named MODEL-CONFIG-FORMAT, e.g.\ LLAMA-ALL-BF16, and is compared against the dense model in the same format.
We test release-format BF16 and llama.cpp's 4-bit Q4\_K\_M to compare the two-bit middle tier against different weight precisions. Most QWEN blocks use Gated DeltaNet linear attention. We evaluate QWEN only in CA, applying \method to its FFN weights while keeping attention weights in GPU memory.

\begin{table}[!htb]
\centering
\caption{The models, configurations and weight formats we evaluate, and their identifiers.}
\vspace{-6pt} %
\label{tab:setup}
\small
\begin{tabularx}{\textwidth}{@{}l l >{\raggedright\arraybackslash}X@{}}
\toprule
	& \textbf{Identifier} & \textbf{Name} \\
\midrule
MODEL & LLAMA & Llama-3.1-8B \citep{grattafiori2024llama3} \\
 & GEMMA & Gemma-4-31B \citep{gemmateam2026gemma4} \\
 & QWEN & Qwen3.6-27B \citep{qwen2026qwen36} \\
\midrule
CONFIG & ALL & FFN and attention projection weights offloaded \\
 & CA & FFN weights offloaded, attention projection weights cached in GPU memory \\
\midrule
FORMAT & BF16 & 16-bit \\
 & Q4KM & Q4\_K\_M \citep{gerganov2023llamacpp}, 4-bit \\
\bottomrule
\end{tabularx}
\vspace{-12pt} %
\end{table}

We compare against prior training-free activation sparsity methods, TEAL \citep{liu2025teal}, WINA \citep{chen2026wina}, LaRoSA \citep{liu2025larosa} and R-Sparse \citep{zhang2025rsparse}.
TEAL selects by magnitude rank rather than against its calibrated threshold, so that it reads exactly the bytes it is given.
Model quality is evaluated using perplexity measured over the WikiText-2 test data split \citep{merity2017pointer} and downstream task accuracy measured using lm-eval \citep{gao2024evalharness} on ARC-Challenge, HellaSwag, PIQA, WinoGrande, MMLU and GSM8K \citep{clark2018arc,zellers2019hellaswag,bisk2020piqa,sakaguchi2020winogrande,hendrycks2021mmlu,cobbe2021gsm8k}.
We calibrate hyperparameters of \method ($k_{top}, k_{mid}$, see Section~\ref{sec:method:calibration}) and all prior works we compare with using WikiText-2 train data split (disjoint from the test split).
Full details are in Appendix~\ref{app:setup}.

\vspace{-6pt} %

\subsection{Byte-Accuracy Trade-off}
\label{sec:results:accuracy}

\vspace{-6pt} %

\begin{figure}[!htb]
\centering
\captionsetup[subfigure]{skip=2pt}
\captionsetup{skip=4pt}
\begin{subfigure}[t]{0.453\textwidth}
\centering
\includegraphics[width=\linewidth]{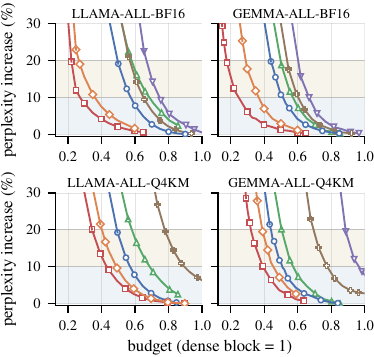}
\caption{FFN and attention projection offloaded}
\label{fig:ppl:all}
\end{subfigure}\hfill
\begin{subfigure}[t]{0.522\textwidth}
\centering
\includegraphics[width=\linewidth]{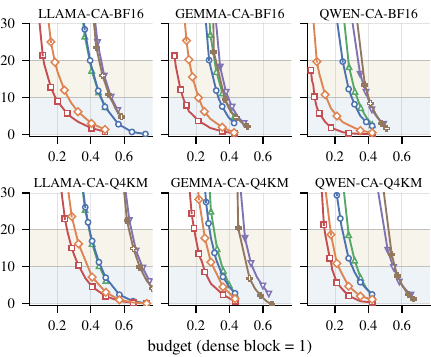}
\caption{FFN offloaded, attention projection cached}
\label{fig:ppl:ca}
\end{subfigure}

\vspace{0pt}
\includegraphics[width=0.886\textwidth]{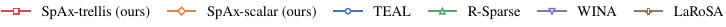}
\vspace{-6pt}
\caption{Perplexity increase over dense against the budget on WikiText-2 test.
The shaded bands mark the two \emph{tolerances}, a perplexity increase of  $10\%$ and $20\%$.}
\label{fig:ppl}
\vspace{-12pt} %
\end{figure}

Figure~\ref{fig:ppl} plots the perplexity increase against the budget for \method and the four baselines.
We observe that, first, at both perplexity tolerances, \method requires a smaller budget than every baseline in all evaluated cases.
SpAx-trellis's budget advantage persists across the plotted range and narrows near zero perplexity loss.
For LLAMA-ALL-BF16, \method requires a $27\%$ budget at the $10\%$ tolerance, compared with $57\%$ for TEAL.
At the $20\%$ tolerance, the corresponding budgets are $22\%$ and $50\%$.
Second, SpAx-trellis requires a smaller budget than SpAx-scalar at the same perplexity tolerance.
At the $10\%$ tolerance, its budget is $7$ percentage points lower on average (up to $12$).
This is because trellis coding incurs less quantization error than scalar coding at the same two-bit rate (Section~\ref{sec:method:codes}).
Third, \method reduces the budget relative to TEAL by less in Q4\_K\_M than in BF16.
At the $10\%$ tolerance, \method's budget is $15$ percentage points below TEAL's on LLAMA-ALL-Q4KM.
On LLAMA-ALL-BF16, the difference is $30$ points.
This is because the middle-tier codes remain at two bits per weight in both formats, $1/8$ of a BF16 weight but about $2/5$ of a Q4\_K\_M weight, so replacing weights by codes saves a smaller fraction of bytes in Q4\_K\_M.
Fourth, \method requires a smaller budget than TEAL even when sparsity is applied only to the FFN (CA).
On LLAMA-CA-BF16 at the $10\%$ tolerance, \method requires a $19\%$ budget, against $46\%$ for TEAL.

Table~\ref{tab:downstream} reports the downstream accuracy at the two tolerances.
At the $10\%$ tolerance, \method loses up to about $4$ points relative to dense, and about $6$ points on GSM8K.
At the $20\%$ tolerance, the accuracy loss is about twice as large.
At either tolerance, \method achieves accuracy comparable to TEAL while requiring less weight data per token.

\begin{table}[!htb]
\vspace{-8pt} %
\centering
\caption{Downstream accuracy of LLAMA-ALL-BF16 for \method and TEAL, the closest baseline in Figure~\ref{fig:ppl}, at the two tolerances.
Appendix~\ref{app:downstream} gives all models and settings.}
\vspace{-8pt} %
\label{tab:downstream}
\footnotesize
\setlength{\tabcolsep}{3.5pt}
\begin{tabular}{lrrrrrrrr}
\toprule
 & Budget & PPL & Arc-C & HS & PIQA & WG & MMLU & GSM8K \\
\midrule
LLAMA-BF16 (Dense) & 100\% & 6.24 & 52.1 & 60.8 & 79.1 & 75.1 & 65.4 & 50.2 \\
\cmidrule(lr){2-9}
TEAL-ALL $+$10\% & 57\% & 6.89\,{\small$(+10.4\%)$} & 48.4 & 58.5 & 78.8 & 72.3 & 62.3 & 42.2 \\
\textbf{SpAx-ALL $+$10\%} & \textbf{27\%} & 6.87\,{\small$(+10.1\%)$} & 49.1 & 58.5 & 78.6 & 73.4 & 62.1 & 43.9 \\
TEAL-ALL $+$20\% & 50\% & 7.44\,{\small$(+19.2\%)$} & 47.4 & 56.7 & 78.0 & 71.0 & 59.1 & 36.0 \\
\textbf{SpAx-ALL $+$20\%} & \textbf{22\%} & 7.47\,{\small$(+19.7\%)$} & 48.1 & 56.9 & 77.4 & 70.6 & 60.1 & 39.2 \\
\bottomrule
\end{tabular}

\vspace{-10pt} %
\end{table}

\vspace{-4pt} %
\subsection{Decoding Speeds}
\label{sec:results:speed}
\label{sec:results:dram}

\begin{figure}[!htb]
\centering
\captionsetup[subfigure]{skip=2pt}
\captionsetup{skip=4pt}
\begin{subfigure}[t]{0.401\textwidth}
\centering
\includegraphics[width=\linewidth]{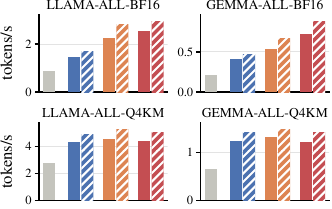}
\caption{CPU memory, ALL}
\label{fig:speed:dram:all}
\end{subfigure}\hfill
\begin{subfigure}[t]{0.564\textwidth}
\centering
\includegraphics[width=\linewidth]{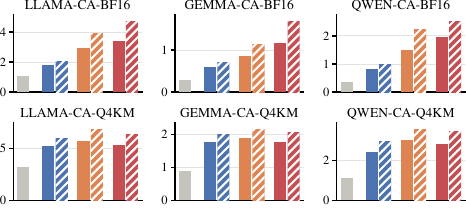}
\caption{CPU memory, CA}
\label{fig:speed:dram:ca}
\end{subfigure}

\vspace{2pt}
\begin{subfigure}[t]{0.412\textwidth}
\centering
\includegraphics[width=\linewidth]{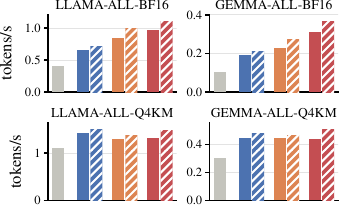}
\caption{Flash, ALL}
\label{fig:speed:flash:all}
\end{subfigure}\hfill
\begin{subfigure}[t]{0.564\textwidth}
\centering
\includegraphics[width=\linewidth]{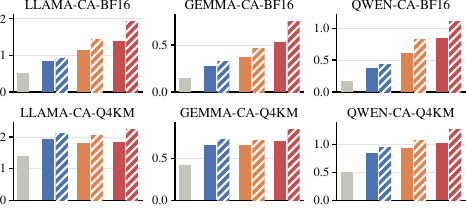}
\caption{Flash, CA}
\label{fig:speed:flash:ca}
\end{subfigure}

\vspace{0pt}
\includegraphics[width=0.820\textwidth]{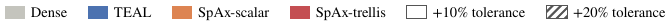}
\vspace{-3pt}
\caption{Decoding speed with the weights offloaded to CPU memory (a, b) and to flash storage (c, d), in the ALL and CA configurations of Table~\ref{tab:setup}.
Each method runs at its budget for each tolerance.}
\label{fig:speed}
\vspace{-9pt} %
\end{figure}

\vspace{-8pt} %
\paragraph{CPU memory offloading}
Figure~\ref{fig:speed}(a,\,b) reports the decoding speed of the dense model, TEAL, SpAx-trellis, and SpAx-scalar.
First, we observe that with BF16 weights, SpAx-trellis decodes fastest across all evaluated models and configurations (up to $2.60\times$, $2.07\times$ on average over TEAL, and up to $7.10\times$, $4.42\times$ faster on average than dense model).
Second, with Q4\_K\_M weights, SpAx-scalar decodes fastest across the evaluated models and configurations (up to $1.25\times$, $1.10\times$ on average over TEAL) although SpAx-trellis reads fewer bytes.
This is because of shorter GPU reconstruction times of SpAx-scalar (Table~\ref{tab:profile}).
Third, caching the attention projections in GPU memory speeds up \method more than TEAL.
From LLAMA-ALL-BF16 to LLAMA-CA-BF16, SpAx-trellis becomes $3.23$--$4.44\times$ faster than dense, while TEAL becomes $1.69$--$1.95\times$ over dense.
This is because \method saves more bytes on the FFN weights compared to attention projections.

\vspace{-8pt} %
\paragraph{Flash offloading}
Figure~\ref{fig:speed}(c,\,d) reports the decoding speed of the same methods.
First, as with CPU memory offloading, SpAx-trellis decodes fastest with BF16 weights across all evaluated models and configurations.
It is $1.92\times$ faster than TEAL and $3.78\times$ faster than dense on average (up to $2.57\times$ and $6.22\times$).
Second, unlike with CPU memory offloading, SpAx-trellis outperforms SpAx-scalar in most Q4\_K\_M cases.
Because reading from flash is slower, the bytes SpAx-trellis saves now outweigh its longer GPU reconstruction.
Third, with Q4\_K\_M weights, SpAx-trellis provides a smaller speedup over TEAL ($1.07\times$ on average).
This is because flash reads have a coarser granularity than the reads of \method (Table~\ref{tab:profile}).

\paragraph{Decoding time breakdown}
\label{sec:results:analysis}

Table~\ref{tab:profile} shows the breakdown of decoding time into transfer time, GPU computation (including middle-tier reconstruction), and the rest.
We observe that, first, SpAx-trellis spends more GPU time ($52$--$64$~ms) than SpAx-scalar ($22$--$27$~ms) due to longer reconstruction times (Section~\ref{sec:method:codes}).
In Q4\_K\_M with CPU memory offloading, SpAx-trellis transfers fewer bytes than SpAx-scalar and spends $24$~ms less on the transfer, but its reconstruction takes $30$~ms longer on the GPU. 
Second, unlike CPU memory offloading, flash offloading transfers more bytes than a token needs: about $1.52\times$ in Q4\_K\_M. This is due to fragmentation: flash reads have a coarse granularity, while \method's column ranges are often small, especially in Q4\_K\_M, where columns are smaller. Even so, SpAx-trellis transfers fewer bytes than TEAL in all settings.

\begin{table}[!htb]
\vspace{-8pt} %
\centering
\caption{Breakdown of decoding time per token of LLAMA-ALL in both formats with CPU memory and flash offloading.
Appendix~\ref{app:profile} gives all models and settings.}
\vspace{-10pt} %
\label{tab:profile}
\footnotesize
\setlength{\tabcolsep}{2pt}
\renewcommand{\arraystretch}{0.9}
\begin{tabular}{lrr rrr rrrr}
\toprule
 & & & \multicolumn{3}{c}{CPU memory (ms)} & \multicolumn{4}{c}{Flash (ms)} \\
\cmidrule(lr){4-6} \cmidrule(lr){7-10}
 & needed (GB) & GPU (ms) & transfer & other & total & moved (GB) & transfer & other & total \\
\midrule
BF16 (Dense) & 13.96 & 26 & 1,118 & 12 & 1,156 & 13.96 & 2,388 & 60 & 2,474 \\
TEAL $+$10\% & 7.95 & 14 & 637 & 29 & 680 & 7.99 & 1,382 & 142 & 1,538 \\
SpAx-trellis $+$10\% & 3.76 & 64 & 302 & 31 & 397 & 4.11 & 834 & 134 & 1,032 \\
SpAx-scalar $+$10\% & 4.83 & 27 & 388 & 34 & 449 & 5.25 & 1,016 & 146 & 1,190 \\
\midrule
Q4KM (Dense) & 4.18 & 18 & 336 & 12 & 366 & 4.18 & 825 & 53 & 896 \\
TEAL $+$10\% & 2.36 & 11 & 191 & 29 & 231 & 3.12 & 587 & 104 & 702 \\
SpAx-trellis $+$10\% & 1.76 & 52 & 142 & 33 & 227 & 2.67 & 588 & 116 & 756 \\
SpAx-scalar $+$10\% & 2.05 & 22 & 166 & 34 & 222 & 3.13 & 646 & 106 & 773 \\
\bottomrule
\end{tabular}
\vspace{-5pt} %
\end{table}

\vspace{-3pt} %
\paragraph{Comparison to weight quantization}
\label{sec:results:quantization}
\vspace{-3pt} %

\begin{figure}[!htb]
\vspace{-6pt} %
\centering
\begin{minipage}[t]{0.48\textwidth}
\centering
\includegraphics[width=\linewidth]{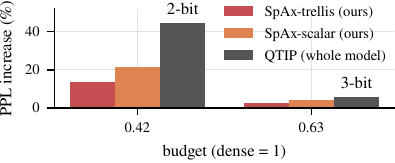}
\vspace{-8pt}
\caption{Perplexity increase at the budgets of 2- and 3-bit QTIP, on LLAMA-ALL-Q4KM.}
\label{fig:quantize}
\end{minipage}\hfill
\begin{minipage}[t]{0.48\textwidth}
\centering
\includegraphics[width=\linewidth]{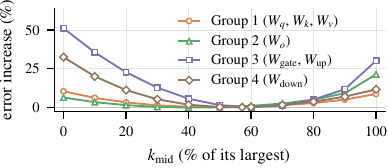}
\vspace{-8pt}
\caption{Layer output error of each $k_{\mathrm{mid}}$ for layer 15 of LLAMA-ALL-Q4KM.}
\label{fig:tier-ratio}
\end{minipage}
\vspace{-12pt} %
\end{figure}

We compare \method against Llama-3.1-8B quantized to 2 and 3 bits with QTIP~\citep{tseng2024qtip} without fine-tuning (Figure~\ref{fig:quantize}).
We find that \method reaches a lower perplexity than QTIP at both budgets.
At the budget of 2-bit QTIP, QTIP raises perplexity by $44.8\%$, against $13.5\%$ for SpAx-trellis and $21.5\%$ for SpAx-scalar.
This is because QTIP lowers the precision of every column, while \method keeps the columns of the largest activation entries in the deployment's weight format (Section~\ref{sec:motivation}).

\subsection{Sensitivity to the Size of the Middle Tier}
\label{sec:results:ablation}
\vspace{-6pt} %

At a fixed budget, we sweep each weight matrix group's $k_{\mathrm{mid}}$ grid on layer 15 of LLAMA-ALL-Q4KM, with the other groups fixed (Figure~\ref{fig:tier-ratio}).
First, both ends of the grid leave more error than the $k_{\mathrm{mid}}$ \method chooses.
Setting $k_{\mathrm{mid}}=0$ raises the layer's error by $6$--$51\%$, and the largest $k_{\mathrm{mid}}$ raises it by $9$--$30\%$.
This is because a larger $k_{\mathrm{mid}}$ recovers more of the omitted contributions but leaves a smaller $k_{\mathrm{top}}$.
Second, the best $k_{\mathrm{mid}}$ depends on the weight format.
It lies at $50$--$60\%$ of the largest $k_{\mathrm{mid}}$ in Q4\_K\_M and at $70$--$86\%$ in BF16. 
Appendix~\ref{app:fractions} provides further ablations. For every weight group of a layer, we sweep $k_{\mathrm{mid}}$ at the group's fixed budget and report the tier sizes and layer error at each grid point, the group's share of the layer budget, and the tier sizes selected by Step~2 (Figure~\ref{fig:overview}).

\vspace{-6pt}
\section{Conclusion}
\label{sec:conclusion}
\vspace{-6pt}

We present \method, a training-free mechanism to accelerate autoregressive decoding in LLMs whose weights exceed GPU memory and must be fetched from CPU memory or flash storage. \method reduces the weight data required per token by skipping weight columns associated with low-magnitude activations or reading compressed approximations of those columns.
Across state-of-the-art LLMs, \method requires less weight data per token than prior training-free activation-sparsity baselines at evaluated perplexity loss tolerances. At a WikiText-2 perplexity increase of at most $10\%$, with weights offloaded to CPU memory, it achieves decoding speedups over dense inference of $3.86\times$ on average (up to $5.57\times$) in BF16 and $2.06\times$ (up to $2.74\times$) with 4-bit weights. With flash offloading, the speedups are $3.31\times$ (up to $4.81\times$) and $1.54\times$ (up to $2.03\times$).

\bibliography{iclr2026_conference}
\bibliographystyle{iclr2026_conference}

\ifincludesupplement
\clearpage
\appendix

{\Huge \textbf{Appendix}}
\bigskip

\section{Experimental Setup Details}
\label{app:setup}

\textbf{Baselines.} We compare against four training-free activation sparsity methods: TEAL \citep{liu2025teal}, WINA \citep{chen2026wina}, LaRoSA \citep{liu2025larosa} and R-Sparse \citep{zhang2025rsparse}.
TEAL selects by magnitude rank rather than against its calibrated threshold, so that it reads exactly the bytes it is given.

\textbf{Metrics.} We report the \emph{read budget}: the weight data required for one decoded token of a block, expressed in bytes and normalized by the block's dense weight size.
\method's codes, two bits per weight, are counted in those bytes, the codebook they index included: 256 KB for the whole model.
The rotations WINA and LaRoSA read and R-Sparse's low-rank factors are counted too, at 16 bits in both weight formats, which is the precision their releases store them at.

\textbf{Evaluation.} Perplexity is measured over the whole WikiText-2 test split \citep{merity2017pointer} in non-overlapping 2048-token windows, and downstream accuracy by lm-eval \citep{gao2024evalharness} on ARC-Challenge \citep{clark2018arc}, HellaSwag \citep{zellers2019hellaswag}, PIQA \citep{bisk2020piqa} and WinoGrande \citep{sakaguchi2020winogrande} 0-shot, and on MMLU \citep{hendrycks2021mmlu} and GSM8K \citep{cobbe2021gsm8k} 5-shot, each benchmark in full.
Every method is applied to the whole of each scored sequence, except on GSM8K: the model generates its answer there, so the prompt is read densely and the method applies to the generated tokens.
The two offline searches of \method (Section~\ref{sec:method:calibration}) and the calibration of every baseline that needs one use the WikiText-2 train split, which is disjoint from the test split.

\textbf{Offloading setup.} We implement \method in llama.cpp and decode at batch one, with every projection computed on the GPU.
The embeddings, the output head and the KV cache stay in GPU memory, as do the attention projection weights in CA.
In CPU memory offloading, the offloaded weights and \method's codes are held in pinned CPU memory, and each decoding step transfers the selected ranges into GPU memory over PCIe.
In flash offloading, they are read from flash storage (the NVMe SSD) with direct I/O, so that every read is served by the drive rather than by the page cache.
Both settings run on one workstation with an RTX 4090 on a PCIe 4.0 x8 link, 128 GB of CPU memory and an NVMe SSD.
A prompt is read densely, and the tiered read applies to the decoding steps.
Each reported latency is the median over eight decoding steps, taken after four warm-up steps, and the bytes those steps read are metered on the same run.

\section{\texorpdfstring{$k_{\mathrm{top}}$ and $k_{\mathrm{mid}}$ at One Layer}{k-top and k-mid at One Layer}}
\label{app:fractions}

Table~\ref{tab:tiers} gives the $k_{\mathrm{top}}$ and $k_{\mathrm{mid}}$ that the two searches of Section~\ref{sec:method:calibration} choose for layer 15 of Llama-3.1-8B.
Here the budget covers the seven weight matrices of one layer.
Every layer runs at a budget of $24.6\%$ ($107.4$ of $436.2$ MB) in BF16 and about $39\%$ ($54.0$ of $138.9$ MB) in Q4\_K\_M.
At these budgets, the perplexity of the model is $12.0\%$ above dense in BF16 and $13.5\%$ in Q4\_K\_M.
Beside each group of matrices is the share of the layer's budget that Step~1 gives it.
The top, middle and bottom columns give $k_{\mathrm{top}}/n$, $k_{\mathrm{mid}}/n$ and the rest, in percent.
Each row changes the $k_{\mathrm{mid}}$ of one group and holds the other groups at the $k_{\mathrm{mid}}$ Step~2 chose, shown in bold.
Error is the layer's output error against reading every weight, on calibration data.

\begin{table}[!ht]
\centering
\caption{$k_{\mathrm{top}}$ and $k_{\mathrm{mid}}$ of each group at layer 15 of Llama-3.1-8B.}
\label{tab:tiers}
\small
\setlength{\tabcolsep}{5pt}
\begin{subtable}[t]{0.49\linewidth}
\centering
\caption{BF16}
\label{tab:tiers-bf16}
\begin{tabular}{lrrrr}
\toprule
 & top & middle & bottom & error \\
\midrule
$W_q$, $W_k$, $W_v$ & 30.0 & 0.0 & 70.0 & 0.1619 \\
$14.0\%$ & 29.0 & 8.0 & 63.0 & 0.1465 \\
($15.1$ MB) & 28.0 & 16.0 & 56.0 & 0.1361 \\
 & 27.0 & 24.0 & 49.0 & 0.1292 \\
 & 26.0 & 32.0 & 42.0 & 0.1248 \\
 & 25.0 & 40.0 & 35.0 & 0.1220 \\
 & 24.0 & 48.0 & 28.0 & 0.1205 \\
 & \textbf{23.0} & \textbf{56.0} & \textbf{21.0} & \textbf{0.1197} \\
 & 22.0 & 64.0 & 14.0 & 0.1195 \\
 & 21.0 & 72.0 & 7.0 & 0.1196 \\
 & 20.0 & 80.0 & 0.0 & 0.1200 \\
\midrule
$W_o$ & 30.0 & 0.0 & 70.0 & 0.1438 \\
$9.4\%$ & 29.0 & 8.0 & 63.0 & 0.1346 \\
($10.1$ MB) & 28.0 & 16.0 & 56.0 & 0.1285 \\
 & 27.0 & 24.0 & 49.0 & 0.1244 \\
 & 26.0 & 32.0 & 42.0 & 0.1219 \\
 & 25.0 & 40.0 & 35.0 & 0.1205 \\
 & 24.0 & 48.0 & 28.0 & 0.1198 \\
 & 23.0 & 56.0 & 21.0 & 0.1197 \\
 & \textbf{22.0} & \textbf{64.0} & \textbf{14.0} & \textbf{0.1197} \\
 & 21.0 & 72.0 & 7.0 & 0.1200 \\
 & 20.0 & 80.0 & 0.0 & 0.1204 \\
\midrule
$W_{\mathrm{gate}}$, $W_{\mathrm{up}}$ & 25.7 & 0.0 & 74.3 & 0.2837 \\
$56.3\%$ & 24.7 & 8.5 & 66.9 & 0.2417 \\
($60.4$ MB) & 23.6 & 17.0 & 59.4 & 0.2071 \\
 & 22.5 & 25.5 & 52.0 & 0.1786 \\
 & 21.5 & 34.0 & 44.6 & 0.1564 \\
 & 20.4 & 42.4 & 37.1 & 0.1400 \\
 & 19.3 & 50.9 & 29.7 & 0.1290 \\
 & 18.3 & 59.4 & 22.3 & 0.1226 \\
 & 17.2 & 67.9 & 14.9 & 0.1200 \\
 & \textbf{16.6} & \textbf{72.8} & \textbf{10.6} & \textbf{0.1197} \\
 & 16.2 & 76.4 & 7.4 & 0.1200 \\
 & 15.1 & 84.9 & 0.0 & 0.1212 \\
\midrule
$W_{\mathrm{down}}$ & 18.6 & 0.0 & 81.4 & 0.2216 \\
$20.3\%$ & 17.4 & 9.3 & 73.3 & 0.1861 \\
($21.8$ MB) & 16.2 & 18.6 & 65.1 & 0.1606 \\
 & 15.1 & 27.9 & 57.0 & 0.1429 \\
 & 13.9 & 37.2 & 48.9 & 0.1312 \\
 & 12.8 & 46.5 & 40.7 & 0.1242 \\
 & 11.6 & 55.8 & 32.6 & 0.1207 \\
 & 10.4 & 65.1 & 24.4 & 0.1197 \\
 & \textbf{10.3} & \textbf{66.5} & \textbf{23.3} & \textbf{0.1197} \\
 & 9.3 & 74.4 & 16.3 & 0.1203 \\
 & 8.1 & 83.8 & 8.1 & 0.1217 \\
 & 6.9 & 93.1 & 0.0 & 0.1236 \\
\bottomrule
\end{tabular}

\end{subtable}\hfill
\begin{subtable}[t]{0.49\linewidth}
\centering
\caption{Q4\_K\_M}
\label{tab:tiers-q4km}
\begin{tabular}{lrrrr}
\toprule
 & top & middle & bottom & error \\
\midrule
$W_q$, $W_k$, $W_v$ & 43.3 & 0.0 & 56.7 & 0.1408 \\
$12.2\%$ & 39.3 & 9.7 & 51.0 & 0.1354 \\
($6.6$ MB) & 35.4 & 19.3 & 45.3 & 0.1317 \\
 & 31.4 & 29.0 & 39.7 & 0.1293 \\
 & 27.4 & 38.6 & 34.0 & 0.1280 \\
 & \textbf{23.4} & \textbf{48.3} & \textbf{28.3} & \textbf{0.1277} \\
 & 19.4 & 57.9 & 22.7 & 0.1282 \\
 & 15.4 & 67.6 & 17.0 & 0.1294 \\
 & 11.5 & 77.2 & 11.3 & 0.1311 \\
 & 7.5 & 86.9 & 5.7 & 0.1340 \\
 & 3.5 & 96.5 & 0.0 & 0.1386 \\
\midrule
$W_o$ & 40.0 & 0.0 & 60.0 & 0.1357 \\
$7.0\%$ & 36.0 & 9.0 & 55.0 & 0.1319 \\
($3.8$ MB) & 32.0 & 18.0 & 50.0 & 0.1294 \\
 & 28.0 & 27.0 & 45.0 & 0.1279 \\
 & 24.0 & 36.0 & 40.0 & 0.1274 \\
 & \textbf{20.0} & \textbf{45.0} & \textbf{35.0} & \textbf{0.1277} \\
 & 16.0 & 54.0 & 30.0 & 0.1288 \\
 & 12.0 & 63.0 & 25.0 & 0.1308 \\
 & 8.0 & 72.0 & 20.0 & 0.1339 \\
 & 4.0 & 81.0 & 15.0 & 0.1393 \\
 & 0.0 & 90.0 & 10.0 & 0.1549 \\
\midrule
$W_{\mathrm{gate}}$, $W_{\mathrm{up}}$ & 42.1 & 0.0 & 57.9 & 0.1929 \\
$51.5\%$ & 37.9 & 9.5 & 52.6 & 0.1730 \\
($27.8$ MB) & 33.7 & 19.0 & 47.3 & 0.1566 \\
 & 29.5 & 28.4 & 42.1 & 0.1439 \\
 & 25.3 & 37.9 & 36.8 & 0.1347 \\
 & 21.1 & 47.4 & 31.5 & 0.1294 \\
 & \textbf{16.9} & \textbf{56.9} & \textbf{26.2} & \textbf{0.1277} \\
 & 12.6 & 66.4 & 21.0 & 0.1291 \\
 & 8.4 & 75.9 & 15.7 & 0.1340 \\
 & 4.2 & 85.3 & 10.4 & 0.1426 \\
 & 0.0 & 94.8 & 5.2 & 0.1660 \\
 & & & & \\
\midrule
$W_{\mathrm{down}}$ & 32.9 & 0.0 & 67.1 & 0.1691 \\
$29.3\%$ & 29.9 & 9.7 & 60.4 & 0.1531 \\
($15.8$ MB) & 27.0 & 19.3 & 53.7 & 0.1419 \\
 & 24.0 & 29.0 & 47.0 & 0.1342 \\
 & 21.1 & 38.6 & 40.3 & 0.1298 \\
 & 18.1 & 48.3 & 33.6 & 0.1278 \\
 & \textbf{16.0} & \textbf{55.2} & \textbf{28.8} & \textbf{0.1277} \\
 & 15.2 & 57.9 & 26.9 & 0.1279 \\
 & 12.3 & 67.6 & 20.1 & 0.1294 \\
 & 9.3 & 77.3 & 13.4 & 0.1323 \\
 & 6.4 & 86.9 & 6.7 & 0.1365 \\
 & 3.4 & 96.6 & 0.0 & 0.1424 \\
\bottomrule
\end{tabular}

\end{subtable}
\end{table}

\clearpage

\clearpage
\section{Decoding Time Breakdown of Llama, Gemma and Qwen}
\label{app:profile}

Tables~\ref{tab:profile-llama}, \ref{tab:profile-gemma} and~\ref{tab:profile-qwen} give the breakdown of Table~\ref{tab:profile} for Llama-3.1-8B in every configuration and weight format, Gemma-4-31B and Qwen3.6-27B.
Qwen is evaluated in CA only.

\begin{table}[!ht]
\centering
\caption{Breakdown of the decoding time per token of Llama-3.1-8B with CPU memory and flash offloading.}
\label{tab:profile-llama}
\small
\setlength{\tabcolsep}{2pt}
\begin{tabular}{llrr rrr rrrr}
\toprule
 & & & & \multicolumn{3}{c}{CPU memory} & \multicolumn{4}{c}{Flash} \\
\cmidrule(lr){5-7} \cmidrule(lr){8-11}
 & & needed & GPU & transfer & other & total & moved & transfer & other & total \\
 & & (GB) & (ms) & (ms) & (ms) & (ms) & (GB) & (ms) & (ms) & (ms) \\
\midrule
LLAMA-ALL-BF16 & Dense & 13.96 & 26 & 1,118 & 12 & 1,156 & 13.96 & 2,388 & 60 & 2,474 \\
 & TEAL $+$10\% & 7.95 & 14 & 637 & 29 & 680 & 7.99 & 1,382 & 142 & 1,538 \\
 & SpAx-trellis $+$10\% & 3.76 & 64 & 302 & 31 & 397 & 4.11 & 834 & 134 & 1,032 \\
 & SpAx-scalar $+$10\% & 4.83 & 27 & 388 & 34 & 449 & 5.25 & 1,016 & 146 & 1,190 \\
\cmidrule(lr){2-11}
LLAMA-CA-BF16 & Dense & 11.27 & 31 & 903 & 11 & 945 & 11.27 & 1,844 & 60 & 1,935 \\
 & TEAL $+$10\% & 6.44 & 22 & 516 & 22 & 560 & 6.44 & 1,066 & 105 & 1,192 \\
 & SpAx-trellis $+$10\% & 2.60 & 61 & 209 & 23 & 293 & 2.85 & 550 & 100 & 712 \\
 & SpAx-scalar $+$10\% & 3.62 & 28 & 290 & 26 & 344 & 3.92 & 730 & 122 & 880 \\
\cmidrule(lr){2-11}
LLAMA-ALL-Q4KM & Dense & 4.18 & 18 & 336 & 12 & 366 & 4.18 & 825 & 53 & 896 \\
 & TEAL $+$10\% & 2.36 & 11 & 191 & 29 & 231 & 3.12 & 587 & 104 & 702 \\
 & SpAx-trellis $+$10\% & 1.76 & 52 & 142 & 33 & 227 & 2.67 & 588 & 116 & 756 \\
 & SpAx-scalar $+$10\% & 2.05 & 22 & 166 & 34 & 222 & 3.13 & 646 & 106 & 773 \\
\cmidrule(lr){2-11}
LLAMA-CA-Q4KM & Dense & 3.41 & 23 & 274 & 11 & 308 & 3.41 & 650 & 39 & 712 \\
 & TEAL $+$10\% & 1.88 & 18 & 152 & 21 & 191 & 2.39 & 422 & 78 & 518 \\
 & SpAx-trellis $+$10\% & 1.37 & 53 & 110 & 25 & 188 & 1.94 & 408 & 80 & 540 \\
 & SpAx-scalar $+$10\% & 1.60 & 21 & 129 & 25 & 175 & 2.33 & 462 & 68 & 550 \\
\bottomrule
\end{tabular}

\end{table}

\begin{table}[!ht]
\centering
\caption{Breakdown of the decoding time per token of Gemma-4-31B with CPU memory and flash offloading.}
\label{tab:profile-gemma}
\small
\setlength{\tabcolsep}{1.8pt}
\begin{tabular}{llrr rrr rrrr}
\toprule
 & & & & \multicolumn{3}{c}{CPU memory} & \multicolumn{4}{c}{Flash} \\
\cmidrule(lr){5-7} \cmidrule(lr){8-11}
 & & needed & GPU & transfer & other & total & moved & transfer & other & total \\
 & & (GB) & (ms) & (ms) & (ms) & (ms) & (GB) & (ms) & (ms) & (ms) \\
\midrule
GEMMA-ALL-BF16 & Dense & 58.57 & 103 & 4,690 & 34 & 4,827 & 58.57 & 9,389 & 170 & 9,662 \\
 & TEAL $+$10\% & 29.20 & 44 & 2,339 & 74 & 2,456 & 30.55 & 4,868 & 343 & 5,255 \\
 & SpAx-trellis $+$10\% & 13.30 & 226 & 1,067 & 84 & 1,377 & 15.18 & 2,631 & 374 & 3,231 \\
 & SpAx-scalar $+$10\% & 21.40 & 76 & 1,716 & 88 & 1,880 & 23.77 & 3,913 & 396 & 4,385 \\
\cmidrule(lr){2-11}
GEMMA-CA-BF16 & Dense & 41.62 & 103 & 3,331 & 30 & 3,464 & 41.62 & 6,576 & 140 & 6,818 \\
 & TEAL $+$10\% & 19.61 & 63 & 1,571 & 52 & 1,686 & 20.61 & 3,252 & 242 & 3,557 \\
 & SpAx-trellis $+$10\% & 7.11 & 222 & 570 & 59 & 851 & 8.18 & 1,392 & 256 & 1,870 \\
 & SpAx-scalar $+$10\% & 12.54 & 84 & 1,006 & 64 & 1,154 & 14.19 & 2,334 & 290 & 2,709 \\
\cmidrule(lr){2-11}
GEMMA-ALL-Q4KM & Dense & 17.51 & 75 & 1,403 & 32 & 1,510 & 17.51 & 3,110 & 144 & 3,330 \\
 & TEAL $+$10\% & 8.74 & 30 & 708 & 72 & 810 & 10.96 & 1,936 & 288 & 2,254 \\
 & SpAx-trellis $+$10\% & 6.81 & 181 & 550 & 86 & 816 & 9.38 & 1,853 & 263 & 2,297 \\
 & SpAx-scalar $+$10\% & 7.68 & 52 & 621 & 82 & 755 & 10.61 & 2,029 & 180 & 2,261 \\
\cmidrule(lr){2-11}
GEMMA-CA-Q4KM & Dense & 12.60 & 90 & 1,009 & 29 & 1,128 & 12.60 & 2,180 & 130 & 2,400 \\
 & TEAL $+$10\% & 5.69 & 52 & 458 & 50 & 561 & 7.06 & 1,271 & 207 & 1,530 \\
 & SpAx-trellis $+$10\% & 4.17 & 173 & 335 & 59 & 567 & 5.81 & 1,096 & 150 & 1,420 \\
 & SpAx-scalar $+$10\% & 5.05 & 64 & 407 & 59 & 530 & 7.09 & 1,360 & 100 & 1,524 \\
\bottomrule
\end{tabular}

\end{table}

\begin{table}[!ht]
\centering
\caption{Breakdown of decoding time per token of Qwen3.6-27B with CPU memory and flash offloading.}
\label{tab:profile-qwen}
\small
\setlength{\tabcolsep}{1.8pt}
\begin{tabular}{llrr rrr rrrr}
\toprule
 & & & & \multicolumn{3}{c}{CPU memory} & \multicolumn{4}{c}{Flash} \\
\cmidrule(lr){5-7} \cmidrule(lr){8-11}
 & & needed & GPU & transfer & other & total & moved & transfer & other & total \\
 & & (GB) & (ms) & (ms) & (ms) & (ms) & (GB) & (ms) & (ms) & (ms) \\
\midrule
QWEN-CA-BF16 & Dense & 34.23 & 58 & 2,739 & 35 & 2,832 & 34.23 & 5,485 & 109 & 5,652 \\
 & TEAL $+$10\% & 14.41 & 25 & 1,156 & 56 & 1,236 & 14.90 & 2,452 & 205 & 2,682 \\
 & SpAx-trellis $+$10\% & 3.78 & 145 & 303 & 60 & 508 & 4.62 & 836 & 193 & 1,174 \\
 & SpAx-scalar $+$10\% & 6.87 & 44 & 551 & 65 & 660 & 7.79 & 1,374 & 230 & 1,648 \\
\cmidrule(lr){2-11}
QWEN-CA-Q4KM & Dense & 10.36 & 42 & 830 & 25 & 897 & 10.36 & 1,848 & 87 & 1,978 \\
 & TEAL $+$10\% & 4.36 & 15 & 352 & 44 & 411 & 5.60 & 1,038 & 121 & 1,174 \\
 & SpAx-trellis $+$10\% & 2.60 & 100 & 209 & 46 & 355 & 3.76 & 747 & 126 & 973 \\
 & SpAx-scalar $+$10\% & 3.12 & 26 & 251 & 50 & 328 & 4.57 & 925 & 112 & 1,063 \\
\bottomrule
\end{tabular}

\end{table}

\clearpage
\section{Downstream Accuracy of Llama, Gemma and Qwen}
\label{app:downstream}

Tables~\ref{tab:downstream-llama}, \ref{tab:downstream-gemma} and~\ref{tab:downstream-qwen} give the downstream accuracy of Table~\ref{tab:downstream} for Llama-3.1-8B, Gemma-4-31B and Qwen3.6-27B.

\begin{table}[!htb]
\centering
\caption{Downstream accuracy of Llama-3.1-8B for \method and TEAL at the two tolerances.}
\label{tab:downstream-llama}
\small
\setlength{\tabcolsep}{3.5pt}
\begin{tabular}{lrrrrrrrr}
\toprule
 & Budget & PPL & Arc-C & HS & PIQA & WG & MMLU & GSM8K \\
\midrule
LLAMA-BF16 (Dense) & 100\% & 6.24 & 52.1 & 60.8 & 79.1 & 75.1 & 65.4 & 50.2 \\
\cmidrule(lr){2-9}
TEAL-ALL $+$10\% & 57\% & 6.89\,{\small$(+10.4\%)$} & 48.4 & 58.5 & 78.8 & 72.3 & 62.3 & 42.2 \\
\textbf{SpAx-ALL $+$10\%} & \textbf{27\%} & 6.87\,{\small$(+10.1\%)$} & 49.1 & 58.5 & 78.6 & 73.4 & 62.1 & 43.9 \\
TEAL-ALL $+$20\% & 50\% & 7.44\,{\small$(+19.2\%)$} & 47.4 & 56.7 & 78.0 & 71.0 & 59.1 & 36.0 \\
\textbf{SpAx-ALL $+$20\%} & \textbf{22\%} & 7.47\,{\small$(+19.7\%)$} & 48.1 & 56.9 & 77.4 & 70.6 & 60.1 & 39.2 \\
\cmidrule(lr){2-9}
TEAL-CA $+$10\% & 46\% & 6.86\,{\small$(+9.9\%)$} & 49.8 & 59.5 & 78.2 & 71.0 & 63.3 & 44.0 \\
\textbf{SpAx-CA $+$10\%} & \textbf{19\%} & 6.88\,{\small$(+10.2\%)$} & 48.5 & 59.3 & 78.8 & 71.9 & 62.7 & 44.7 \\
TEAL-CA $+$20\% & 40\% & 7.49\,{\small$(+20.0\%)$} & 47.5 & 57.7 & 78.6 & 70.6 & 60.5 & 36.4 \\
\textbf{SpAx-CA $+$20\%} & \textbf{11\%} & 7.58\,{\small$(+21.4\%)$} & 47.1 & 58.0 & 78.5 & 71.5 & 60.1 & 37.5 \\
\midrule
LLAMA-Q4KM (Dense) & 100\% & 6.44 & 51.4 & 60.0 & 78.7 & 73.6 & 64.7 & 48.6 \\
\cmidrule(lr){2-9}
TEAL-ALL $+$10\% & 57\% & 7.11\,{\small$(+10.4\%)$} & 48.5 & 57.6 & 77.9 & 71.1 & 60.9 & 37.9 \\
\textbf{SpAx-ALL $+$10\%} & \textbf{42\%} & 7.12\,{\small$(+10.6\%)$} & 48.3 & 57.9 & 78.9 & 72.2 & 60.5 & 39.7 \\
TEAL-ALL $+$20\% & 50\% & 7.68\,{\small$(+19.3\%)$} & 45.0 & 55.8 & 76.9 & 71.7 & 56.8 & 32.8 \\
\textbf{SpAx-ALL $+$20\%} & \textbf{34\%} & 7.73\,{\small$(+20.1\%)$} & 46.0 & 56.3 & 77.6 & 70.6 & 56.7 & 32.8 \\
\cmidrule(lr){2-9}
TEAL-CA $+$10\% & 45\% & 7.15\,{\small$(+11.1\%)$} & 49.5 & 58.8 & 77.6 & 72.5 & 61.8 & 39.7 \\
\textbf{SpAx-CA $+$10\%} & \textbf{33\%} & 7.10\,{\small$(+10.2\%)$} & 50.0 & 59.4 & 78.3 & 72.5 & 61.9 & 39.6 \\
TEAL-CA $+$20\% & 39\% & 7.72\,{\small$(+19.9\%)$} & 47.8 & 58.0 & 77.4 & 71.9 & 59.6 & 34.6 \\
\textbf{SpAx-CA $+$20\%} & \textbf{25\%} & 7.79\,{\small$(+21.0\%)$} & 48.3 & 57.9 & 77.8 & 70.0 & 59.0 & 35.7 \\
\bottomrule
\end{tabular}
\end{table}

\begin{table}[!htb]
\centering
\caption{Downstream accuracy of Gemma-4-31B for \method and TEAL at the two tolerances.}
\label{tab:downstream-gemma}
\small
\setlength{\tabcolsep}{3.5pt}
\begin{tabular}{lrrrrrrr}
\toprule
 & Budget & PPL & Arc-C & HS & PIQA & WG & MMLU \\
\midrule
GEMMA-BF16 (Dense) & 100\% & 6.39 & 64.8 & 65.8 & 82.9 & 80.1 & 84.5 \\
\cmidrule(lr){2-8}
TEAL-ALL $+$10\% & 50\% & 7.07\,{\small$(+10.7\%)$} & 63.3 & 62.8 & 82.2 & 77.7 & 82.4 \\
\textbf{SpAx-ALL $+$10\%} & \textbf{23\%} & 7.12\,{\small$(+11.5\%)$} & 63.0 & 64.0 & 81.2 & 79.1 & 82.8 \\
TEAL-ALL $+$20\% & 44\% & 7.65\,{\small$(+19.8\%)$} & 59.2 & 60.8 & 81.6 & 76.5 & 80.6 \\
\textbf{SpAx-ALL $+$20\%} & \textbf{18\%} & 7.63\,{\small$(+19.4\%)$} & 61.2 & 63.2 & 81.5 & 76.2 & 82.0 \\
\midrule
GEMMA-Q4KM (Dense) & 100\% & 6.54 & 63.7 & 65.7 & 82.8 & 79.6 & 83.6 \\
\cmidrule(lr){2-8}
TEAL-ALL $+$10\% & 50\% & 7.26\,{\small$(+11.1\%)$} & 62.0 & 62.9 & 82.2 & 76.9 & 81.6 \\
\textbf{SpAx-ALL $+$10\%} & \textbf{39\%} & 7.20\,{\small$(+10.1\%)$} & 62.7 & 63.9 & 82.3 & 77.4 & 82.3 \\
TEAL-ALL $+$20\% & 43\% & 7.94\,{\small$(+21.5\%)$} & 59.5 & 60.8 & 80.8 & 75.4 & 80.1 \\
\textbf{SpAx-ALL $+$20\%} & \textbf{32\%} & 7.82\,{\small$(+19.7\%)$} & 61.3 & 62.7 & 81.9 & 75.1 & 80.7 \\
\bottomrule
\end{tabular}

\end{table}

\begin{table}[!htb]
\centering
\caption{Downstream accuracy of Qwen3.6-27B for \method and TEAL at the two tolerances.}
\label{tab:downstream-qwen}
\small
\setlength{\tabcolsep}{3.5pt}
\begin{tabular}{lrrrrrr}
\toprule
 & Budget & PPL & Arc-C & HS & PIQA & WG \\
\midrule
QWEN-BF16 (Dense) & 100\% & 7.40 & 60.3 & 64.0 & 81.6 & 77.2 \\
\cmidrule(lr){2-7}
TEAL-CA $+$10\% & 30\% & 8.18\,{\small$(+10.6\%)$} & 58.7 & 62.2 & 80.9 & 75.3 \\
\textbf{SpAx-CA $+$10\%} & \textbf{8\%} & 8.16\,{\small$(+10.2\%)$} & 57.6 & 62.9 & 80.5 & 74.3 \\
TEAL-CA $+$20\% & 25\% & 8.86\,{\small$(+19.7\%)$} & 56.3 & 60.2 & 79.3 & 73.5 \\
\textbf{SpAx-CA $+$20\%} & \textbf{6\%} & 8.69\,{\small$(+17.4\%)$} & 56.4 & 62.2 & 80.3 & 73.2 \\
\midrule
QWEN-Q4KM (Dense) & 100\% & 7.47 & 59.0 & 63.9 & 81.4 & 77.0 \\
\cmidrule(lr){2-7}
TEAL-CA $+$10\% & 30\% & 8.27\,{\small$(+10.7\%)$} & 57.3 & 62.4 & 80.1 & 75.5 \\
\textbf{SpAx-CA $+$10\%} & \textbf{18\%} & 8.32\,{\small$(+11.4\%)$} & 58.3 & 62.0 & 80.2 & 75.6 \\
TEAL-CA $+$20\% & 24\% & 9.04\,{\small$(+20.9\%)$} & 56.4 & 60.6 & 79.8 & 73.6 \\
\textbf{SpAx-CA $+$20\%} & \textbf{13\%} & 9.07\,{\small$(+21.3\%)$} & 56.7 & 60.4 & 79.6 & 74.3 \\
\bottomrule
\end{tabular}

\end{table}

\fi

\end{document}